\documentclass[conference]{IEEEtran}
\IEEEoverridecommandlockouts
\usepackage{fancyhdr}
\usepackage{booktabs}
\usepackage{tabularx}
\usepackage{float}
\usepackage{pdflscape}
\usepackage{cite}
\usepackage{amsmath,amssymb,amsfonts}
\usepackage{algorithmic}
\usepackage{graphicx}
\usepackage{textcomp}
\usepackage{xcolor}
\usepackage{comment}
\usepackage{subcaption}
\usepackage{soul}

\def\BibTeX{{\rm B\kern-.05em{\sc i\kern-.025em b}\kern-.08em
    T\kern-.1667em\lower.7ex\hbox{E}\kern-.125emX}}
\begin{document}

\title{An Embedded Real-time Object Alert System for Visually Impaired: A Monocular Depth Estimation based Approach through Computer Vision}

\author{
\IEEEauthorblockN{%
    \begin{minipage}{0.4\textwidth}
        \centering
        Jareen Anjom\\
        Department of Electrical \& Computer Engineering\\
        \textit{North South University}\\
        Dhaka, Bangladesh\\
        jareen.anjom@northsouth.edu
    \end{minipage}%
    \hfill
    \begin{minipage}{0.4\textwidth}
        \centering
        Rashik Iram Chowdhury\\
        Department of Electrical \& Computer Engineering\\
        \textit{North South University}\\
        Dhaka, Bangladesh\\
        rashik.chowdhury@northsouth.edu
    \end{minipage}%
}

\vspace{0.5cm}

\IEEEauthorblockN{%
    \begin{minipage}{0.4\textwidth}
        \centering
        Tarbia Hasan\\
        Department of Electrical \& Computer Engineering\\
        \textit{North South University}\\
        Dhaka, Bangladesh\\
        tarbia.hasan@northsouth.edu
    \end{minipage}%
    \hfill
    \begin{minipage}{0.4\textwidth}
        \centering
        Md. Ishan Arefin Hossain\\
        Department of Electrical \& Computer Engineering\\
        \textit{North South University}\\
        Dhaka, Bangladesh\\
        ishan.hossain@northsouth.edu
    \end{minipage}%
}
}

\maketitle

\begin{abstract}
Visually impaired people face significant challenges in their day-to-day commutes in the urban cities of Bangladesh due to the vast number of obstructions on every path. With many injuries taking place through road accidents on a daily basis, it is paramount for a system to be developed that can alert the visually impaired of objects in close distance beforehand. To overcome this issue, a novel alert system is proposed in this research to assist the visually impaired in commuting through these busy streets without colliding with any objects. The proposed system can alert the individual of objects that are present at a close distance. It utilizes transfer learning to train models for depth estimation and object detection and combines both models to introduce a novel system. The models are optimized through the utilization of quantization techniques to make them lightweight and efficient, allowing them to be easily deployed on embedded systems. The proposed solution achieved a lightweight real-time depth estimation and object detection model with an mAP50 of 0.801.
\end{abstract}


\section{Introduction}
Visually impaired people face significant challenges in their day-to-day commutes in the urban cities of Bangladesh due to the vast number of obstructions on every path. With many injuries occurring through road accidents daily, it is paramount for a system to be developed that can alert the visually impaired of objects at close distance beforehand. Thus, a novel alert system is proposed in this research to assist the visually impaired in commuting through these busy streets without colliding with any objects. The proposed system can alert individuals to objects that are present at a close distance. It utilizes transfer learning to train models for depth estimation and object detection and combines both models to introduce a novel system. The models are optimized through quantization techniques to make them lightweight and efficient, allowing them to be easily deployed on embedded systems. The proposed solution achieved a lightweight real-time depth estimation and object detection model with a mAP50 of 0.801.

The significant rise in the number of accidents on the streets of Bangladesh is a growing concern for the safety of pedestrians. A largely affected group of people is the visually impaired, who find it difficult to move across the busy streets without accidentally injuring themselves from various obstructions such as motorcycles, cars, pedestrians, rickshaw pullers, etc. This problem is primarily relevant in countries like Bangladesh, the eighth most densely populated country in the world \cite{ref1}. Urban areas in the capital city of Dhaka face the busiest traffic and pedestrian crowds daily. In this country, about 750 thousand people are blind, and six million more have varying degrees of visual impairment \cite{ref2}. According to a report \cite{ref3}, the number of pedestrian fatalities in Dhaka is comparatively higher than in cities of other countries. With the prevalence of accidents, it is crucial to aid the visually impaired in safely moving through such obstructions. 

The visually impaired are accustomed to using walking sticks or canes to identify if any object is blocking their path ahead of them. Since they cannot easily identify the object type, several solutions have been implemented by researchers, such as the Smart Walker\cite{ref13}, wearable spectacles for object identification \cite{ref12}, and other object detection modules. However, most of these solutions are limited to indoor settings. Systems that have worked with outdoor surroundings \cite{ref9} \cite{ref10} \cite{ref11} have been beneficial for the visually impaired to identify dangerous objects blocking their path on the streets. However, objects on the streets of Bangladesh can be quite different from those in other countries. For this purpose, a Bangladesh Road Scene dataset RSUD20K \cite{ref14} is utilized in this research to help the visually impaired individuals who face various challenges in avoiding objects such as rickshaw vans, auto rickshaws, haulers, trucks, bicycles, etc. Then, a depth estimation model is used to identify objects in close proximity. The system utilizes Raspberry Pi camera modules to efficiently detect the depth of the objects ahead of the individuals and successfully identify the objects that are in close distance to avoid clashes.
The objectives of this research are as follows:

\begin{itemize}
    \item Introduce an embedded system using Raspberry Pi that provides real-time alerts without a distributed network.
    \item Combine both object detection and depth estimation models to introduce a transfer-learning based optimized novel system for object identification in close proximity by utilizing two datasets of different domains and apply quantization for generating a lightweight model.
\end{itemize}

The paper is organized in sections as follows: Section \ref{sec:related-works} explores the existing related literature and their limitations, Section \ref{sec:methodology} is the detailed methodology of the system, Section \ref{sec:results} evaluates various object detection models across different metrics. And finally, the paper is concluded in Section \ref{sec:conclusion}.

\section{Related Works}
\label{sec:related-works}
Recent advancements in technologies for visually impaired individuals and autonomous systems have primarily focused either on object detection or on depth estimation to provide users with spatial awareness and obstacle detection in their surroundings. Very few have done both, and most of these works have not optimized their models for lightweight deployment on resource-constrained devices. This gap leaves a need for more efficient solutions that can achieve both tasks while maintaining a compact size and feasible real-time performance for assistive devices. The works listed below highlight the two objectives- Depth estimation and object detection. Table \ref{tab:literature-comparisn} shows a summary comparison of the related works and critically analyses the existing literature.
\begin{table*}[htbp]
    \centering
    \caption{Comparison of Notable Literature}
    \label{lots-of-texts}
    \small 
    \begin{tabular}{lp{2cm}p{1.5cm}p{2.5cm}p{4cm}p{3cm}@{\hskip 0.5 in}p{3cm}}
    \toprule
       Ref.  & Target User & Dataset & Models & Contribution & Limitation \\
       \midrule
        \cite{ref4} & Visually impaired & MS COCO, Flickr30k & SSD and RNN & Vision Navigator Framework & Outdoor performance degradation\\ \addlinespace 

        \cite{ref5} & UAV navigation, autonomous systems & Mid-Air dataset & U-Net + YOLOv3 + Attention & Unsupervised depth estimation & Limited generalization\\ \addlinespace

        \cite{ref6} & Autonomous vehicle developers & Tartan Drive & FCRN, ResNet-50 & Monocular depth estimation & Outputs Not Sharp like SOTA\\ \addlinespace 

        \cite{ref7} & Autonomous driving, robotics & KITTI stereo & Multi-stage architecture, SGM  & monoResMatch: Multi-scale feature extractor & Stereo setup dependency\\ \addlinespace 

        \cite{ref8} & Autonomous driving, collision avoidance & KITTI, Make3D, CityScapes &  PyD-Net variants & Two lightweight architectures, PyD-Net and PyD-Net2 & Accuracy gap with SOTA\\ \addlinespace 

        \cite{ref16} & Autonomous vehicles, robotics & NYU Depth-V2, KITTI & MobileNetV2, encoder-decoder & Patch-Wise Attention Module, Multi-Scale Feature Fusion & Lacks the accuracy of stereo or multi-view depth estimation\\ \addlinespace 

        \cite{ref9} & Visually impaired & COCO Dataset & YOLOv5, Transfer Learning & YOLOv5 with an audio response system & Limited generalization\\ \addlinespace 

        \cite{ref10} & Visually impaired & Custom (OOD) dataset & YOLOv5,v8 Object detection & New dataset, object detection with auditory feedback & Low FPS\\ \addlinespace 

        \cite{ref11} & Visually impaired & TOC, COCO, TT100K & YOLOv5-8, YOLO-NAS  & Real-time obstacle detection & Limited generalization\\ \addlinespace 

        \cite{ref12} & Visually impaired & Real-time environment images & YOLO, Raspberry Pi & Assistive device, high accuracy & Lighting challenges, hardware dependency\\ \addlinespace 

        \cite{ref13} & Visually impaired & Custom images & YOLOv5, ultrasonic sensors & Smart walker navigation & Light sensitivity, obstacle speed\\ \addlinespace

    \bottomrule
    \end{tabular}
    \label{tab:literature-comparisn}
\end{table*}

\subsection{Depth Estimation}
Vision Navigator \cite{ref4} is a system that assists visually impaired people using a smart-fold cane and sensor-equipped shoes called smart-alert walker. Object detection is performed using SSD-RRN to detect the type of objects. For object detection, the smart-alert walker shoe has ultrasonic sensors mounted over it to find objects over a short-range distance. However, the proposed system may not be feasible for objects that are not sensed from ground level.

In the paper by Pian et al. \cite{ref5}, unsupervised stereo depth estimation has been proposed in unstructured environments. A U-shaped CNN model based on Yolov3 residual structure and attention mechanism is developed and evaluated on the Mid-Air dataset, engendering a score of 0.125 and 8.422 for AbsRel and RMSE, respectively.

The paper \cite{ref6} presents a lightweight depth estimation network for predicting top-down terrain depth maps in autonomous navigation systems. The network architecture is based on a Fully Convolutional Residual Network with ResNet-50 as the backbone for feature extraction. The proposed model is evaluated and trained on the Tartan Drive dataset, achieving 0.02039 BerHu Loss and 0.1895 RMSE for aligned and normalized terrain maps. 

Convolutional Neural Networks (CNNs) are applied to the KITTI Dataset for depth estimation purposes. Several existing works have evaluated custom CNN models on this dataset. This paper \cite{ref16} uses a Patch-Wise Attention Network with an encoder-decoder baseline that uses MobileNetV2 as a backbone network. This algorithm achieves 2.927 RMSE. Another study \cite{ref8} applies PyD-Net2 which is a lightweight pyramidal network with only 1.9 million parameters and an RMSE score of 5.529. F. Tosi et al. \cite{ref7} uses their custom model monoResMatch for depth estimation, which is a monocular Residual Matching network that uses a multi-scale feature extractor in a self-supervised manner. This custom model scores an RMSE of 4.351. However, it is an extensive model with a total of 42.5 million parameters.

\subsection{Object Detection}
In this study \cite{ref9}, an outdoor obstacle detection system is proposed using YOLOv5 for efficient object recognition. Google Text-to-Speech is used for audio feedback to alert the user about the objects. The model is based on 10 outdoor objects based on the MS COCO integrated with a custom dataset. YOLOv5 and RCNN models are compared, where YOLOv5 shows better performance in mAP scores.

The paper \cite{ref10} proposes an intelligent obstacle detection system for detecting objects on sidewalks using YOLOv5 and YOLOv8 models. The system is deployed on Raspberry Pi for real-time feedback. The model is trained on a custom dataset with 22 different types of classes.

An outdoor obstacle detection system focusing on YOLO models is proposed in this paper \cite{ref11}. Seven  YOLO models are trained on datasets such as TOC, COCO, and TT100K, which focus on daily-life objects present on sidewalks. Among these models, YOLOv8 achieved the highest precision of 80\% and a recall of 68\%. However, these datasets do not have prevalent images on Bangladeshi streets.

This paper \cite{ref12} introduces a system that uses wearable spectacles linked to a Raspberry Pi camera for real-time object detection to aid visually impaired individuals. A YOLO algorithm is used for object detection purposes with a 90\% accuracy rate in indoor settings.

Another system, Smart Walker \cite{ref13}, utilizes the Raspberry Pi camera module that detects objects further than 1 meter. An ultrasonic sensor attached to the bottom of the walker helps detect objects of small height. The object detection module is trained on objects used daily in the house. A YOLOv5 model is chosen with an accuracy of 86.25\%.

This study \cite{ref17} used another effective method, which introduces a deep learning-based, lightweight edge solution to detect tactile tiles on the footpath and correctly navigate the user through audio-based directions. It uses an HBFN algorithm to navigate the path accurately. The YOLOv8s model gave the best result, which was trained on a novel footpath dataset created through rigorous dataset collection and preprocessing techniques. Optimization techniques were applied to make the model lightweight and deployed to Android application.

\section{Methodology}
\label{sec:methodology}

\begin{figure}[h]
    \centering
    \includegraphics[width=0.75\linewidth]{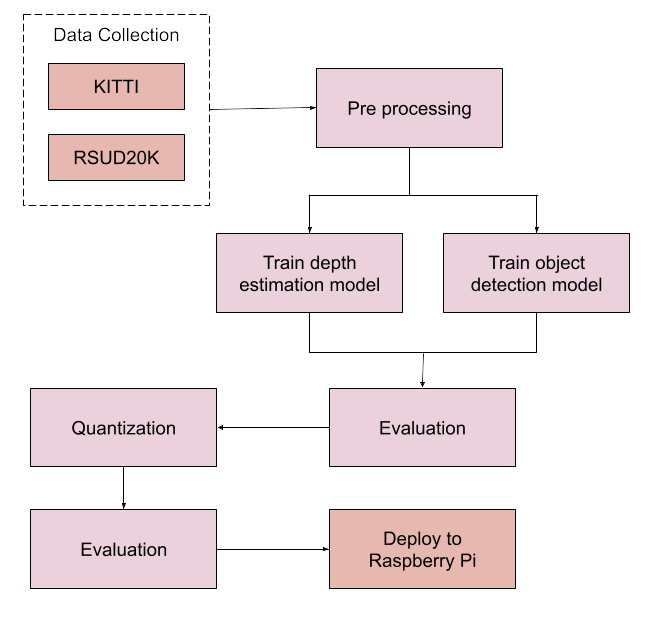}
    \caption{Methodology Flowchart}
    \label{fig:methodology}
\end{figure}

In this research, the steps shown in Fig.~\ref{fig:methodology} are taken to design the alert system for the visually impaired people.

\subsection{Data Collection}
For depth estimation, the KITTI depth dataset is used, which has 86000 training, 7000 validation, and 1000 test set images along with their depth maps.

For detecting local objects on the streets of Bangladesh, a dataset based on the road scenes, RSUD20K \cite{ref14}, is utilized, which is comprised of over 20K high-resolution images and includes 130K bounding box annotations for 13 objects. These objects are person, rickshaw, rickshaw van, auto rickshaw, truck, pickup truck, private car, motorcycle, bicycle, bus, micro bus, covered van, human hauler.

\subsection{Data Preprocessing}
The images of both datasets have been preprocessed by resizing them to a size of 224 by 224. The data is also normalized to [0,1] before passing it for the training of the models. 

\subsection{Training Phase}
In the training phase, transfer learning is incorporated, where the pre-trained weights of the top convolution layers are frozen, and the classifier head is trained entirely on the dataset. This technique leverages the weights learned by the benchmark models on other comprehensive datasets for feature extraction, thus reducing the computational cost.  

For depth estimation, the pre-trained DPT Hybrid MiDaS model is used to predict the depth maps of images. The backbone for this deep learning model architecture uses a combination of ViT (Vision Transformer) and ResNet50. The ViT helps construe long-range dependencies and the image context, while the CNN pre-trained model ResNet50 leverages feature extraction capabilities. The non-negative constraint is set to True for the output depth values representing actual distances. The model is evaluated on the KITTI dataset.

In the case of objection detection, a one-stage detector YOLOv8m is used, which, through transfer learning, is fine-tuned on the RSUD20K \cite{ref14} dataset. The dataset comprises 18,681 images for the train set, 1,004 for the validation set, and 649 for the test set. In order to apply transfer learning of the YOLO models, the Ultralytics library \cite{ref15} is used for training and evaluation. The model is trained for 30 epochs with a batch size 16; the optimizer used is Adam, with a learning rate 0.0001. The pre-trained weights are transferred to train it on the RSUD20K dataset, and the 22nd block of the architecture is frozen during training. In this case, 469 items out of 475 are transferred from the pre-trained weights. This is a key technique in transfer learning where the parameters trained on the original data are copied to the new model.

The loss function components used in the YOLOv8s model are categorized below.

\( L_{\text{box}} \) is the total box loss, which measures the difference between the predicted and actual bounding boxes.
\begin{equation}
L_{\text{box}} = \sum_{i=1}^{N} \left(1 - \text{IoU}(B_i, \hat{B}_i)\right)
\label{eq:box_loss}
\end{equation}

where \( N \) is the total number of bounding boxes, \( B_i \) is the actual bounding box for the \(i\)-th object. \( \hat{B}_i \) is the predicted bounding box for the \(i\)-th object, and \( \text{IoU}(B_i, \hat{B}_i) \) is the Intersection over Union between actual bounding box \(B_i\) and predicted bounding box \(\hat{B}_i\).

\( L_{\text{cls}} \) measures the accuracy of the predicted class probabilities (equation \ref{eq:cls_loss}).

\begin{equation}
L_{\text{cls}} = \sum_{i=1}^{N} \sum_{c=1}^{C} \left[ p_{i,c} \log(\hat{p}_{i,c}) \right]
\label{eq:cls_loss}
\end{equation}
where \( N \) is the total number of bounding boxes, \( C \) is the total number of classes.,\( p_{i,c} \) is the actual probability of the \(i\)-th bounding box belonging to class \(c\), and \( \hat{p}_{i,c} \) is the predicted probability of the \(i\)-th bounding box that belongs to class \(c\).

\( L_{\text{dfl}} \) is the total distribution focal loss that focuses on difficult-to-predict boxes (equation \ref{eq:dfl_loss}).

\begin{equation}
L_{\text{dfl}} = \sum_{i=1}^{N} \text{DFL}(B_i, \hat{B}_i)
\label{eq:dfl_loss}
\end{equation}
where \( N \) is the total number of bounding boxes, \( \text{DFL}(B_i, \hat{B}_i) \) is the distribution focal loss function applied to the actual bounding box \(B_i\) and the predicted bounding box \(\hat{B}_i\).

\begin{equation}
L = \lambda_{\text{box}} L_{\text{box}} + \lambda_{\text{cls}} L_{\text{cls}} + \lambda_{\text{dfl}} L_{\text{dfl}}
\label{eq:combined_loss}
\end{equation}

For evaluation, YOLOv5n, YOLOv5s, and YOLOv8n have been further applied to both respective datasets for comparison.  

\subsection{Quantization}
\label{subsec:quantization}
To make the system more optimized, the proposed models are further quantized to reduce their weights and make them deployable to resource-constraint devices. This is beneficial for deploying such lightweight models to edge devices such as Raspberry Pi, smartphones, etc. 
As shown in Table \ref{tab:quantization}, the depth estimation model is significantly quantized by reducing its weights. For optimization, Tensorflow Lite is used, which reduces its size from 407MB to an impressively compact size of 162.4MB. Similarly, the object detection model is also quantized from 49.6MB to 25MB. 

\begin{table}[htbp]
\caption{Quantization Information}
\begin{center}
\begin{tabular}{|c|c|c|}
\hline
Proposed Model &  Original Size (MB) & Quantized Size (MB)\\
\hline
Depth Estimation &470 & 162.4\\
Object Detection &49.6 & 25\\
\hline

\end{tabular}
\label{tab:quantization}
\end{center}
\end{table}

\subsection{System Architecture}
\begin{figure}[h]
    \centering
    \includegraphics[width=0.80\linewidth]{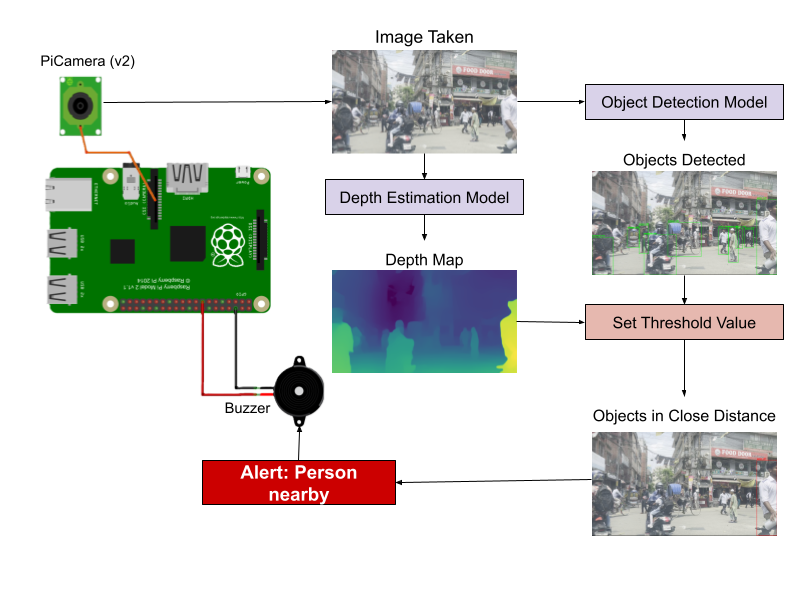}
    \caption{Schematic View of the Proposed System}
    \label{fig:system-schematic}
\end{figure}

The system is designed for an embedded system, Raspberry Pi, which uses the Pi camera to take an image of the scene in front of the visually impaired individual. The image is then passed through the depth estimation model, which generates the depth map scenario. It is also sent to the object detection model to detect and identify objects. A certain threshold is set on the depth map to filter out only the obstructions in close distance that the individual may face. The system then sends out an auditory alert that notifies the individual of the object. The buzzer or auditory alert can be altered according to the user's preference. For example, a vibration method can be added for users who do not like loud sound alerts or audible speech can be delivered to declare the object identified in close proximity. The final system design is shown in Fig.~\ref{fig:system-schematic}.

\section{Results and Evaluation}
\label{sec:results}

In order to evaluate the performance of depth estimation models, AbsRel, SqRel, and RMSE are calculated. 

AbsRel (Absolute Relative) is a measure that takes the average relative difference between the estimated depth and the actual depth \eqref{eq:absrel}. 

\begin{equation}
\label{eq:absrel}
\text{AbsRel} = \frac{1}{N} \sum_{i=1}^{N} \frac{|d_i - \hat{d}_i|}{d_i}
\end{equation}
where $N$ is the total number of pixels, $d_i$ is the estimated depth at pixel $i$, and $\hat{d}_i$ is the actual depth at pixel $i$.

SqRel (Squared Relative) is a measure that takes the average squared relative difference between the estimated depth and actual depth \eqref{eq:sqrel}.

\begin{equation}
\label{eq:sqrel}
\text{SqRel} = \frac{1}{N} \sum_{i=1}^{N} \frac{(d_i - \hat{d}_i)^2}{\hat{d}_i}
\end{equation}

RMSE (Root Mean Squared Error) is a measure of the differences between predicted values by a model and the actual values, i.e., It is the square root of the average of the squared differences between the predicted and actual values \eqref{eq:rmse}. 
\begin{equation}  
\label{eq:rmse}
\text{RMSE} = \sqrt{\frac{1}{n} \sum_{i=1}^{n} (p_i - t_i)^2}
\end{equation}
where \( n \) is the number of samples, \( p_i \) is the predicted value, and \( t_i \) is the true value.

In table \ref{tab:depth-comparison}, the DPT model produced the lowest AbsRel among the three models evaluated on the KITTI dataset.

\begin{table}[htbp]
\caption{Comparison of different depth estimation models}
\begin{center}
\begin{tabular}{|p{3cm}|c|c|c|c|c|c|}
\hline
Model &  AbsRel & SqRel & RMSE & Params (M) \\
\hline
\textbf{DPT Hybrid MiDaS} & \textbf{0.062} & \textbf{0.222}& \textbf{2.575} & \textbf{12.3}\\
MobileNetV2-based + Patch-wise Attention \cite{ref16}&0.069&0.277& 0.2927 &- \\
PyD-Net2 \cite{ref8}&0.127&1.059& 5.259 &1.9 \\
monoResMatch \cite{ref7}& 0.096&0.673&4.351 &42.5 \\

\hline

\end{tabular}
\label{tab:depth-comparison}
\end{center}
\end{table}

Metrics such as precision, recall, F1 score, and mean average precision (mAP) are used to evaluate object detection models. 

Precision measures the accuracy of positive predictions \eqref{eq:precision}.
\begin{equation}
\label{eq:precision}
Precision = \frac{TP}{TP + FP}
\end{equation}
where $TP$ is True Positive and $FP$ is False Positive.

Recall measures the ability to identify all actual instances \eqref{eq:recall}.
\begin{equation}
\label{eq:recall}
Recall = \frac{TP}{TP + FN}
\end{equation}
where $TP$ is True Positive and $FN$ is False Negative.

F1 score balances the trade-off between precision and recall \eqref{eq:f1score}.
\begin{equation}
\label{eq:f1score}
F_1 Score= 2 \cdot \frac{Precision \cdot Recall}{Precision + Recall}
\end{equation}

Mean Average Precision (mAP) combines both precision and recall of multiple classes and different IoU thresholds. The Average Precision of each class is calculated and then averaged over several classes \eqref{eq:map}. 

\begin{equation}
\label{eq:map}
\text{mAP} = \frac{1}{N} \sum_{i=1}^{N} AP_i
\end{equation}
where $N$ is the number of classes, and $AP_i$ is the average precision of class $i$.

In Table \ref{tab:object-comparison}, different YOLO models, along with the current existing work, have been compared to analyze their performance. The best metrics have been highlighted in bold.

\begin{table}[ht]
\caption{Comparison of different object detection models}
\begin{center}
\begin{tabular}{|c|c|c|c|c|}
\hline
Model &  mAP50  & F1 Score & Params (M) & Size (MB)\\
\hline
YOLOv5n &0.706& 0.666 & 2.5 & 5\\
YOLOv5s &0.778  & 0.733 &9.1 &17.6\\
YOLOv8n &0.73  & 0.695 &3 & 5.94\\
\textbf{YOLOv8m} & \textbf{0.801}& \textbf{0.808} &\textbf{25.8} & \textbf{49.6}\\
YOLOv6-M6 \cite{ref14} & 0.779 & - & 79.6 & -\\
\hline

\end{tabular}
\label{tab:object-comparison}
\end{center}
\end{table}

YOLOv8m engendered state-of-the-art performance on the RSUD20K dataset (as seen in table \ref{tab:object-comparison}), surpassing the existing SoTA by 2.82\% on mAP50. These two proposed models are further quantized and evaluated on unseen test datasets. For the depth estimation model, the model has been reduced to 65\% its size while the object detection model has been reduced to approximately 50\% its size. 

After the quantization process, as described in Section \ref{subsec:quantization}, the proposed model engendered 0.769 mAP50  and an F1 score of 0.733. It is observed that the model's performance is not compromised significantly after reducing its weights.

\begin{figure}[h]
    \centering
    \includegraphics[width=0.7\linewidth]{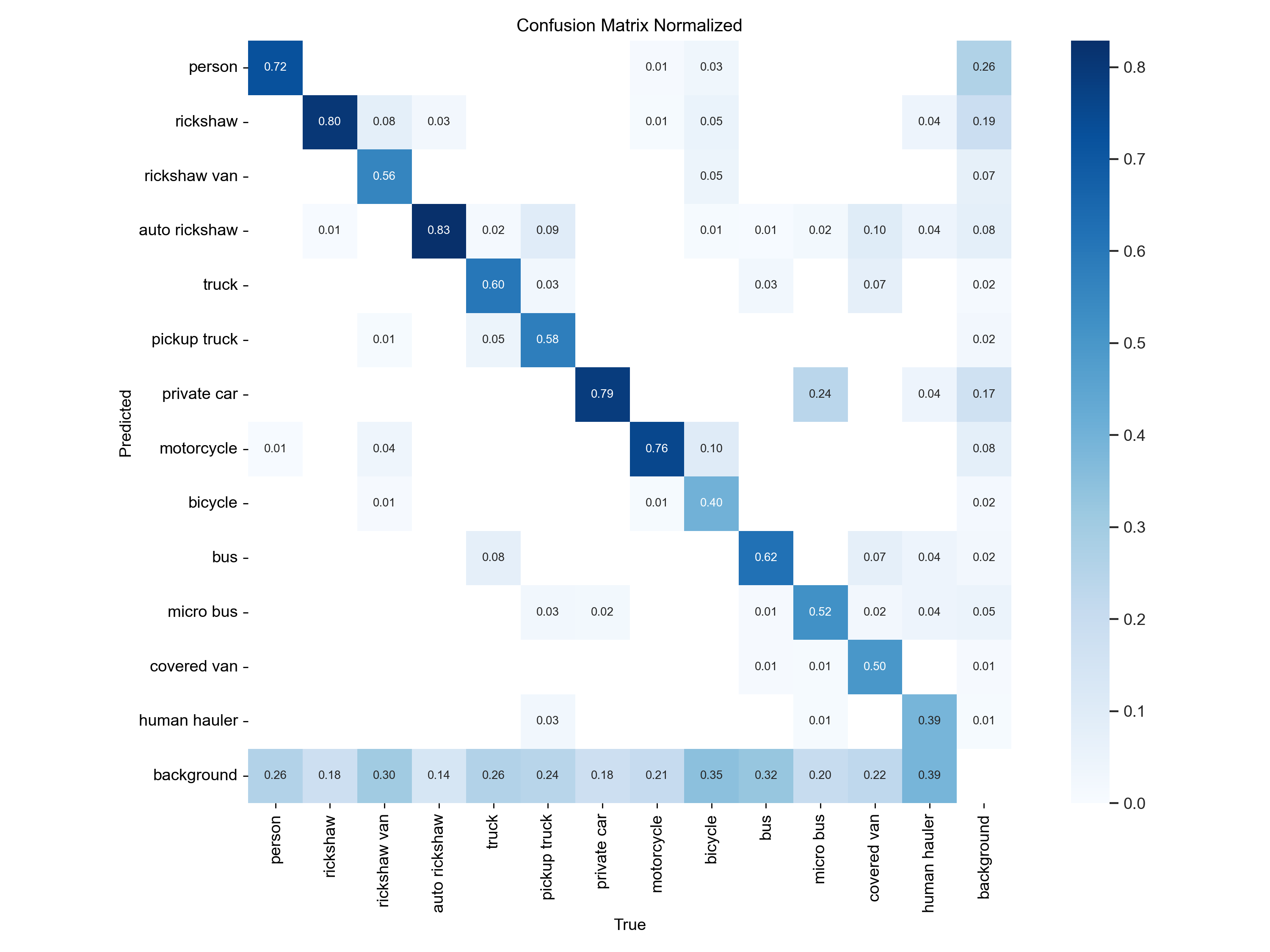}
    \caption{Confusion Matrix}
    \label{fig:odm-confusion-matrix}
\end{figure}

\begin{figure}[h]
    \centering
    \begin{subfigure}[b]{0.48\linewidth}
        \centering
        \includegraphics[width=\linewidth]{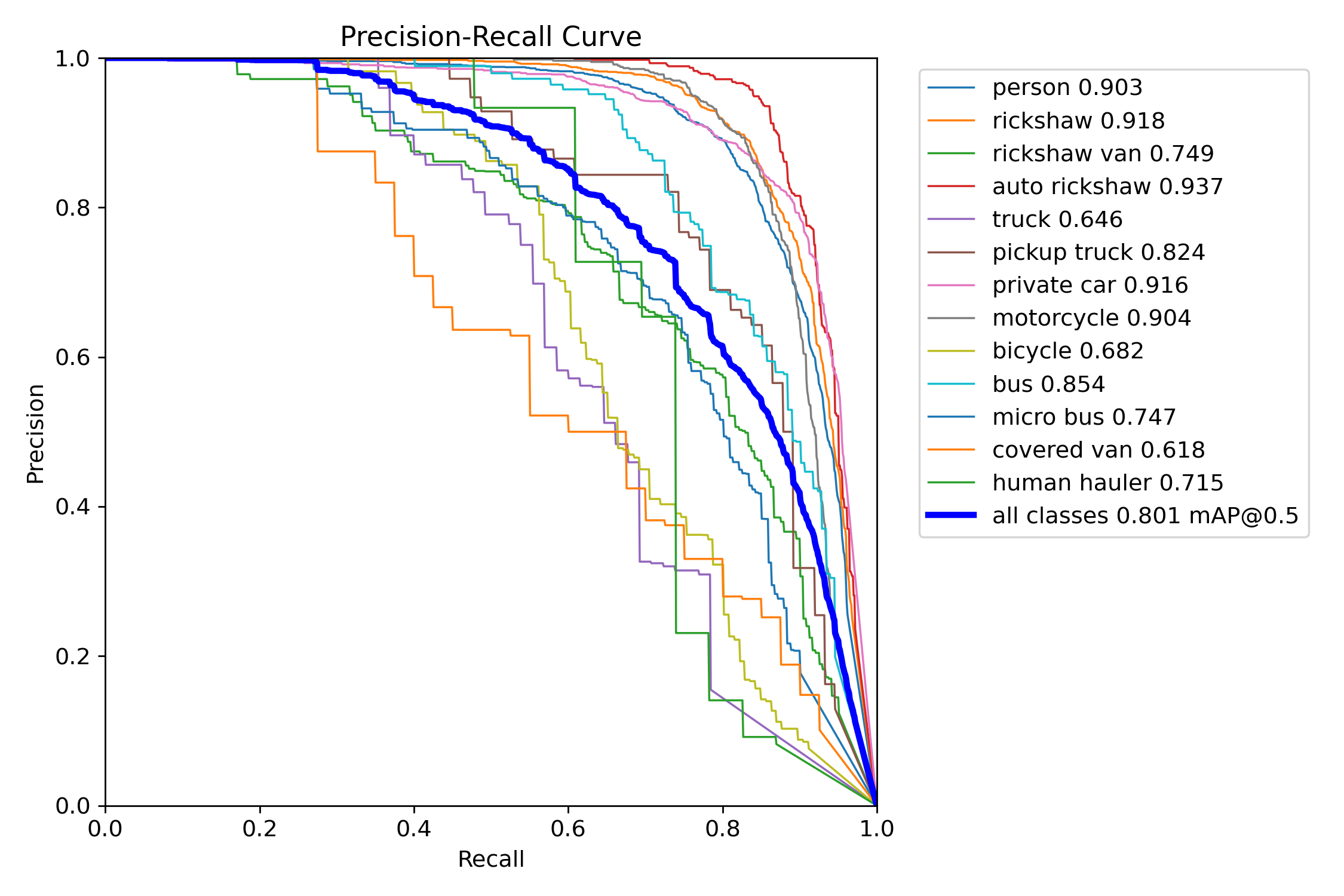}
        \caption{PR Curve}
        \label{fig:odm-pr-curve}
    \end{subfigure}
    \hfill
    \begin{subfigure}[b]{0.48\linewidth}
        \centering
        \includegraphics[width=\linewidth]{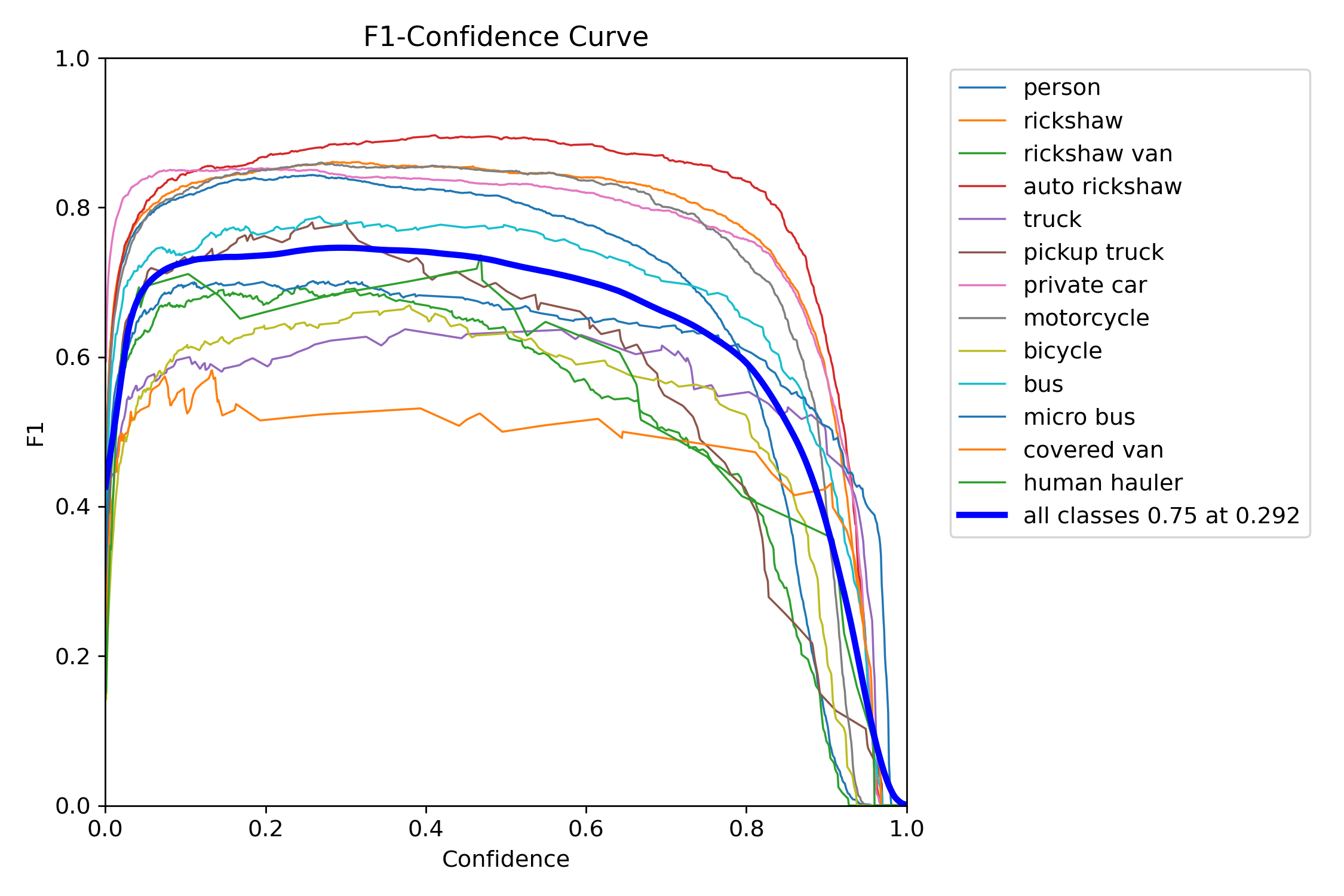}
        \caption{F1 Curve}
        \label{fig:odm-f1-curve}
    \end{subfigure}
    \caption{Performance Curves of the Proposed Object Detection Model}
    \label{fig:odm-curves}
\end{figure}

\begin{figure}[h]
    \centering
    \includegraphics[width=0.8\linewidth]{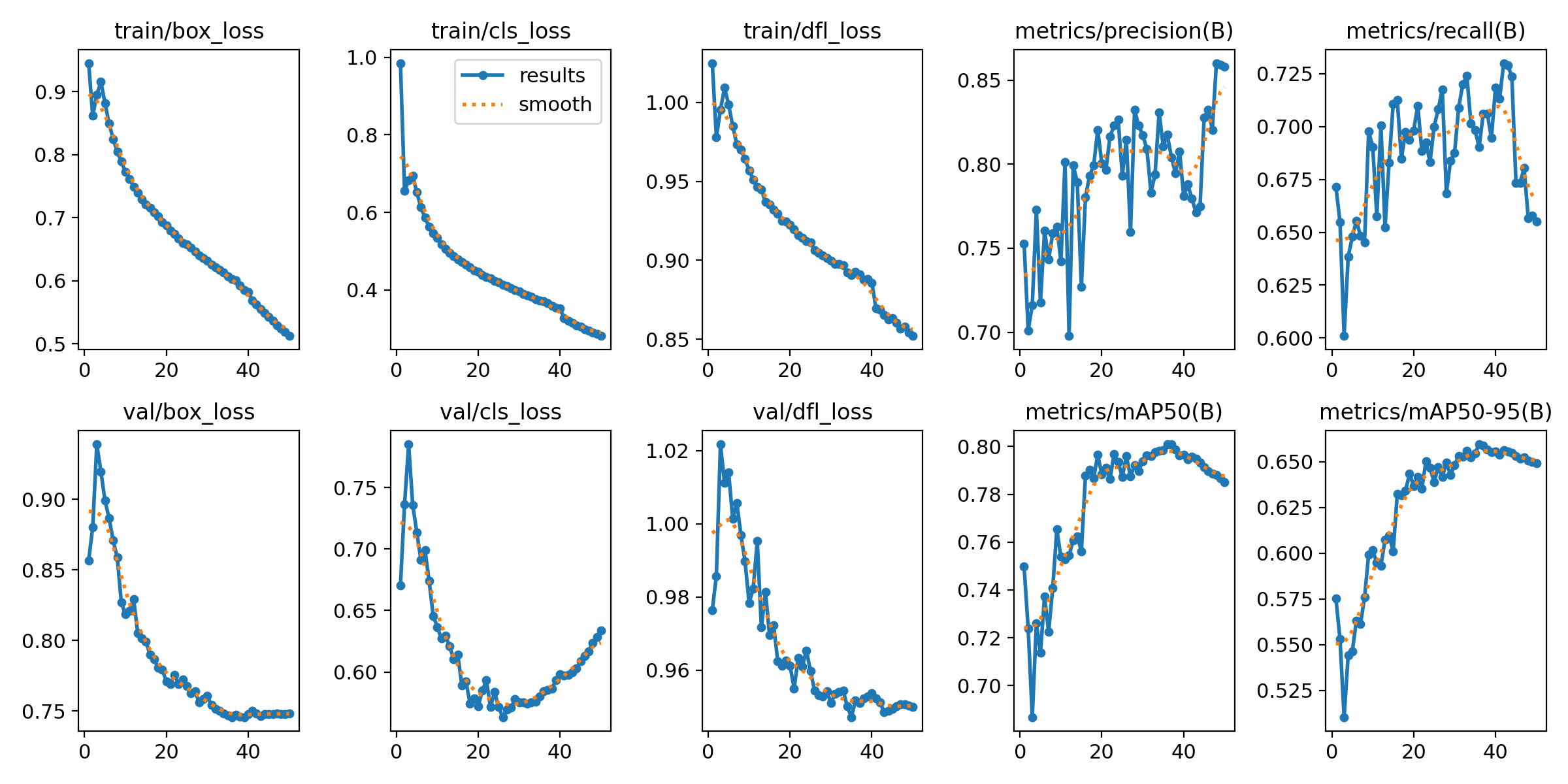}
    \caption{Loss, mAP, Precision, Recall curves}
    \label{fig:results}
\end{figure}

The confusion matrix of the object detection model is shown in Fig.~\ref{fig:odm-confusion-matrix}, PR, and F1 curve in Fig.~\ref{fig:odm-curves}. The curves for box, cls, and dfl loss, precision, recall, mAP50, mAP50-95 are shown in Fig.~\ref{fig:results}.

\section{Conclusion}
\label{sec:conclusion}
Urban environments are often challenging to explore for visually impaired individuals as they face many obstructions throughout their paths that may injure them. It is crucial to understand the type of object that is in their walking vicinity so that they can successfully avoid any clashes. This research proposes an embedded system using Raspberry Pi that can alert the individual of any objects nearby and identify the object type. By leveraging deep edge intelligence techniques, a depth estimation model is used to find the depth of various objects in a scenario. Then by setting a threshold value, an object detection model is used to identify the objects at a close distance. 

The proposed alert system is highly efficient considering its compact size due to quantization. The combination of lightweight depth estimation and object detection models provides a feasible solution for assisting the visually impaired. Both models have achieved a good accuracy as well.

For future work, the proposed system can be further extended by deploying it to other edge devices that are more easily accessible such as smartphones. Explainable AI can be integrated to explore the interpretability of the models, and various other benchmark models can also be applied to improve the system's performance.

\end{document}